\documentclass[sigconf  ]{acmart}

\AtBeginDocument{%
  \providecommand\BibTeX{{%
    \normalfont B\kern-0.5em{\scshape i\kern-0.25em b}\kern-0.8em\TeX}}}

\setcopyright{acmcopyright}
\copyrightyear{2018}
\acmYear{2018}
\acmDOI{XXXXXXX.XXXXXXX}

\acmConference[]{}{}

\usepackage{multirow}
\usepackage{amsmath}

\begin{document}

\title{Adaptive Graph Spatial-Temporal Transformer Network for Traffic Flow Forecasting}

\author{Aosong Feng}
\email{aosong.feng@yale.edu}
\affiliation{%
  \institution{Yale University}
  \city{New Haven}
  \state{Connecticut}
  \country{USA}
}
\author{Leandros Tassiulas}
\email{leandros.tassiulas@yale.edu}
\affiliation{%
  \institution{Yale University}
  \city{New Haven}
  \state{Connecticut}
  \country{USA}
}


\begin{abstract}
Traffic flow forecasting on graphs has real-world applications in many fields, such as transportation system and computer networks. Traffic forecasting can be highly challenging due to complex spatial-temporal correlations and non-linear traffic patterns. Existing works mostly model such spatial-temporal dependencies by considering spatial correlations and temporal correlations separately, and fail to model the direct spatial-temporal correlations. 
Inspired by the recent success of transformers in the graph domain, in this paper, we propose to directly model the cross-spatial-temporal correlations on the spatial-temporal graph using local multi-head self-attentions. To reduce the time complexity, we set the attention receptive field to the spatially neighboring nodes, and we also introduce an adaptive graph to capture the hidden spatial-temporal dependencies. Based on these attention mechanisms, we propose a novel Adaptive Graph Spatial-Temporal Transformer Network (ASTTN), which stacks multiple spatial-temporal attention layers to apply self-attention on the input graph, followed by linear layers for predictions. 
Experimental results on public traffic network datasets, METR-LA PEMS-BAY, PeMSD4, and PeMSD7, demonstrate the superior performance of our model.
\end{abstract}

\ccsdesc[500]{Information systems~Spatial-temporal systems}
\ccsdesc[300]{Information systems~Data mining}
\ccsdesc{Computing methodologies~Neural networks}

\keywords{Traffic prediction, spatial-temporal attention, graph neural networks}

\maketitle

\section{Introduction}
\label{sec:1}
Traffic prediction of spatial-temporal data has attracted much attention in various domains. It predicts the future traffic conditions based on the history records distributed among multiple nodes in the network. This plays an important role in many real-world applications from diverse fields. For example, the predicted future data throughput in the computer network can help network operators to perform real-time traffic steering and routing to improve the user experience.

In recent years, the machine learning research community has put great effort into handling such spatial-temporal data with deep learning model. Convolutional neural networks (CNNs) have been widely used to explore the spatial correlations \cite{shi2020spatial,yao2019revisiting,zhang2017deep}. For temporal modeling, recurrent neural networks are usually deployed to handle the time series \cite{lv2018lc,vinayakumar2017applying}.
The real-world spatial-temporal data usually come with an underlying graph structure, which describes the node correlations using adjacency matrix. Recently, graph modeling on spatial-temporal data has been in the spotlight because of the success of Graph neural networks (GNNs) in the graph domain. Using GNNs (e.g., GCN \cite{kipf2016semi}) to model spatial-temporal correlations especially the spatial correlations has achieved superior performance compared with traditional methods \cite{guo2020optimized,zhao2019t,yu2017spatio,li2017diffusion}. Although significant improvements have been made by incorporating the graph structure into spatial-temporal data forecasting model, current models still face several challenges due to the complexity of spatial-temporal correlations.

\begin{figure}[t]
  \centering \includegraphics[width=0.9\linewidth]{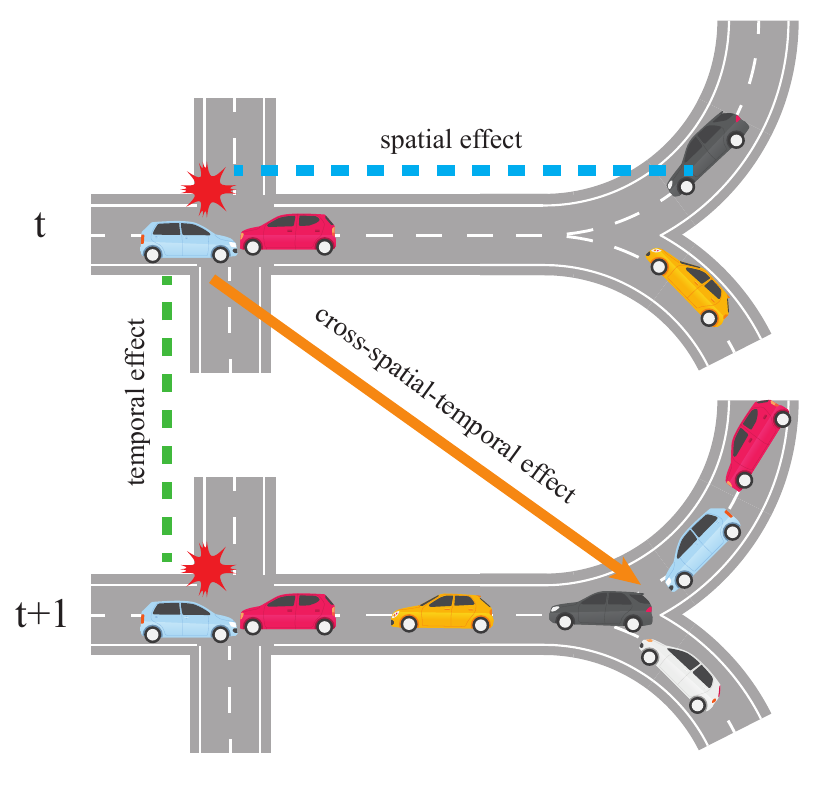}
  \caption{Scenario of spatial-temporal correlations in the road network. Traffic condition at one node can affect others through spatial effect, temporal effect, and cross-spatial-temporal effect.}
  \label{fig:cartoon}
\end{figure}

First, the influence of one node on another node in the spatial graph can span multiple time steps, i.e., the traffic condition of a node at time $t$ can directly affect other nodes at different time step $t'$. For example, as shown in Figure 1, the car accident at the left crossroad at time $t$ is the major cause of the future traffic jam at the right merging lane at time $t+1$. Such spatial-temporal correlations generalize both the spatial correlations and temporal correlations, and are closer to the real causal effect hidden in the spatial-temporal graph-structured data. However, most previous works deal with spatial dimension and temporal dimension separately using different modules and mechanisms, and then fuse the results \cite{li2017diffusion,yan2018spatial}. Such decomposition can be ineffective to model the direct cross-spatial-temporal effect. For example, in Figure 1, the temporal correlations in the right merging lane alone cannot reveal the effect of the car accident, and the spatial correlations at one time step alone do not contain the history traffic dynamics. Therefore, it is important to consider such cross-spatial-temporal effect during the graph modeling.

Second, previous works usually use the predetermined graph structure built from the distance measure or other geographical connections, and use different variants of the adjacency matrix for spatial modeling. These studies are based on the assumption that the graph structure in use can capture the genuine dependency relationships among nodes. However, such geographical connections may not be equivalent to the real traffic correlations. For example, two connected crossroads can have unrelated traffic if no traffic is in the connecting road, on the other hand, the effect of a traffic jam at one crossroad can be broadcast to several hops away, where the roads may not be connected to that crossroad. Therefore, exploring the true spatial dependency is important for spatial modeling. Several works have considered this issue and proposed adaptive graphs \cite{wu2019graph, lu2020spatiotemporal} to explore the hidden correlations, and we follow this idea to address this problem.

Third, even with the graph which captures the true dependencies among nodes, the spatial correlations can be dynamically changing across different time steps. For example, for two connected crossroads, the vehicle traffic during morning and evening rush hours can have opposite directions because of workers commuting between home and office. This indicates we cannot apply the same node updating mechanism at different time steps because the node correlations can also be influenced by the temporal dynamics. How to model such timely-changing spatial correlations and dynamically select relevant nodes' traffic to predict the target traffic remains a challenging issue. Recently, transformer architectures have achieved dominant performance in language modeling \cite{vaswani2017attention,devlin2018bert} and computer vision \cite{dosovitskiy2020image,liu2021swin}. Inspired by the transformer architecture, we propose to use the self-attention mechanism to model the dynamic spatial-temporal correlations in the graph. 

To address the aforementioned challenges, in this work, we propose Adaptive Graph Spatial-Temporal Transformer Network (ASTTN) to collectively predict the traffic flow at every location on the traffic network. Motivated by recent success of applying transformer in modeling video spatial-temporal correlations \cite{arnab2021vivit,liu2021swin}, we perform multi-head self-attention on the spatial-temporal graph and design the ST-attention block to process the graph-structured data. Compared to previous works using separate spatial and temporal modeling, ASTTN only contains stacked ST-attention modules which jointly model the spatial-temporal correlations without decomposing it into spatial and temporal domains. To scale down the time complexity, we specify the construction of the spatial-temporal graph to the 1-hop neighbors in the spatial domain. Besides, we further introduce adaptive spatial graph modeling to explore the genuine correlations for more efficient attention mechanism. The main contributions of this paper are summarized as follows:
\begin{itemize}
    \item We propose to use the local spatial-temporal graph for the spatial-temporal modeling. Specifically, we treat different nodes at different time steps as separate tokens for the transformer input, and limit the attention scope to 1-hop spatial neighbors, which keeps the complexity scalable. 
    \item We utilize the adaptive graph construction to explore the genuine node correlations by selecting nodes that a target node can attend over. This extends attentions beyond 1-hop spatial restriction.
    \item A novel transformer-based network architecture is designed to model diverse types of causal effects in spatial-temporal data. It consist of self-attention-based ST-attention modules to capture the dynamically changing correlations.
    \item Extensive experiments are carried out on real-world high-way traffic datasets, and our model achieves competitive prediction performances compared to the baselines.
\end{itemize}

\section{Related Works}
\subsection{Traffic Forecasting}
Traffic forecasting has been widely explored from different perspectives and found applications in various fields. The conventional statistical models proposed for time-series modeling include HA, ARIMA \cite{williams2003modeling}, VAR \cite{chen2001freeway}. Later, machine learning methods have been applied to traffic prediction such as SVM \cite{jeong2013supervised} and KNN \cite{van2012short}. The rise of deep learning has largely improved the time-series prediction performance and researchers start to handle more complex data by taking both spatial and temporal domains into considerations. Zhang et al. \cite{zhang2017deep} used residual convolution and proposed ST-ResNet to predict the crowd flows. Yao et al. \cite{yao2018deep} used CNN in the spatial domain and long-short term memory (LSTM) in the temporal domain. These modeling methods usually require grid traffic data formulation and fail to consider the non-Euclidean dependencies among nodes. 

\subsection{Machine Learning on Graph}
In recent years, the machine learning research community has devoted substantial energy to applying graph neural networks (GNNs) to numerous downstream graph-related tasks \cite{zhou2018graph}.
Numerous variants of GNNs have been proposed and achieved remarkable results, such as
graph classification \cite{ying2018hierarchical}, node classification \cite{kipf2016semi}, link prediction \cite{zhang2018link}, and community detection \cite{chen2017supervised}.
Recent works have also added the spatial graph formulation into traffic prediction and consider spatial-temporal graph models to handle the graph-structured spatial-temporal data. Most spatial-temporal graph networks follow two directions depending on how to handle the temporal dimension, i.e., RNN-based and CNN-based methods. 
For RNN-based methods, Seo et al. \cite{seo2018structured} proposed to integrate the graph convolution operation into RNN to process the input state and hidden states. DCRNN \cite{li2017diffusion} was later proposed to use diffusion convolution for spatial domain and LSTM for temporal domain.
CNN-based methods combine the graph convolution in the spatial domain and 1-D convolution in the temporal domain, and are shown to be efficient and competitive \cite{yu2017spatio,yan2018spatial}. Both types of methods decompose the spatial-temporal dependencies into spatial and temporal domain. Although such decomposition can be computationally efficient, they have to stack multiple layers to expand the receptive field and cannot address the first problem mentioned in Section \ref{sec:1}.

\subsection{Attention Mechanism}
Attention mechanisms and transformers were initially adopted in the natural language processing and facilitated the large-scale machine learning with model pre-training \cite{devlin2018bert,beltagy2020longformer}. The basic idea of attention mechanism is to dynamically select information that is relevant to the current node from all the other nodes in the input. Recently, researchers have also found successful applications of transformers in the computer vision \cite{dosovitskiy2020image,bao2021beit} and video processing \cite{arnab2021vivit,liu2021swin,bertasius2021space}. Especially, video processing also requires performing attention mechanism on both spatial and temporal domains. TimeSformer \cite{bertasius2021space} considered the direct spatial-temporal attention and used every patch in the video as input. ViViT \cite{arnab2021vivit} factorized the attention to spatial and temporal domain. Video Swin transformer \cite{liu2021swin} utilized the sliding window approach and calculated the local spatial-temporal attention within the window. Our local spatial-temporal model on graph is inspired by these video transformer approach, and we also discuss the influence of different types of attentions in the following sections. 

There are also recent works trying to apply transformers into graph datasets \cite{dwivedi2020generalization,ying2021transformers}, with appropriate structural embeddings and positional embeddings. Our graph spatial-temporal model is also based on such successful practices.
Some works have also applied the attention mechanism into the spatial-temporal modeling. For example, GMAN \cite{zheng2020gman} and STTN \cite{xu2020spatial} added spatial-temporal embedding for each input token, applied spatial attention and temporal attention separately, and then combined the results. However, these approaches cannot directly model the cross-spatial-temporal effect, while our model address this issue by performing spatial-temporal attention jointly.

\section{Methodology}
\subsection{Problem Formulation}
We represent the road network as a weighted graph $G = (\mathcal{V},\mathcal{E},A)$, where $\mathcal{V}$ is the set of nodes with $|\mathcal{V}|=N$ corresponding to $N$ nodes, $\mathcal{E}$ is the set of edges with $|\mathcal{E}|=E$ corresponding to $E$ edges, and $A\in \mathbb{R}^{N\times N}$ is the adjacency matrix describing the spatial distance among nodes. At each time step $t$, the graph will have a feature matrix $X_t\in \mathbb{R}^{N\times D}$ which will be dynamically changing over time $t$. Given a graph $G$ and feature matrix of historical $T$ time steps, the goal of traffic forecasting is to learn a function $f$ which can predict the feature matrix of the future $T'$ time steps. The mapping relationship is shown as follows
\begin{equation}
\label{eq:mapping}
\begin{split}
   [X_{t-T+1:t},G] \xrightarrow{f} X_{t+1:t+T'},
\end{split}
\end{equation}
where $X_{t-T+1:t}\in\mathbb{R}^{T\times N\times D}, X_{t+1:t+T'}\in\mathbb{R}^{T'\times N\times D}$, and we assume $G$ is fixed and not related to the time domain.

\subsection{Multi-Head Self-Attention}
\label{sec:msa}
The key component of the conventional transformer architecture is multi-head self-attention (MSA) which allows a network to attend over all the tokens in the input sequence. Denote input sequence with $N$ tokens and token dimension $D$ as $X\in\mathbb{R}^{N\times D}$. The idea of self-attention is to update each token's own value by querying other tokens  using the corresponding query-key pair. To do this, $X$ is first projected to three matrices: query $Q$, key $K$, and value $V$ as
\begin{equation}
\label{eq:qkv}
\begin{split}
   Q=XW_Q, \hspace{0.5cm} K=XW_K, \hspace{0.5cm} V=XW_V,
\end{split}
\end{equation}
where $W_Q, W_K \in\mathbb{R}^{D\times D_{QK}}$, and $W_V\in\mathbb{R}^{D\times D_V}$. Query $Q$ and key $K$ have the same dimension $D_{QK}$, while $V$ has dimension $D_V$, and in practice, we set $D_{QK}=D_{V}$. Then self-attention can be written as 
\begin{equation}
\label{eq:sa}
\begin{split}
   SA(X) = softmax(\frac{QK^T}{\sqrt{D_{QK}}})V,
\end{split}
\end{equation}
where $softmax$ denotes the row-wise softmax normalization, and we omit the bias term for simplicity. Next, to include multiple aspects that a token wants to attend to, we can further extend Eq. \ref{eq:sa} to multi-head self-attention as follows
\begin{equation}
\label{eq:msa}
\begin{split}
   MSA(X) = concat(SA(X)_1,...,SA(X)_h)W_O,
\end{split}
\end{equation}
where $h$ is the number of heads in use. Translating to the spatial-temporal graph regime, a token in the transformer input sequence refers to a unique node with a location-time pair describing its unique position in the spatial-temporal graph, it can attend over all other nodes at different geographical locations and time steps, or only attend over connected nodes determined by the spatial and temporal adjacency.

\begin{figure*}
  \centering \includegraphics[width=0.95\textwidth]{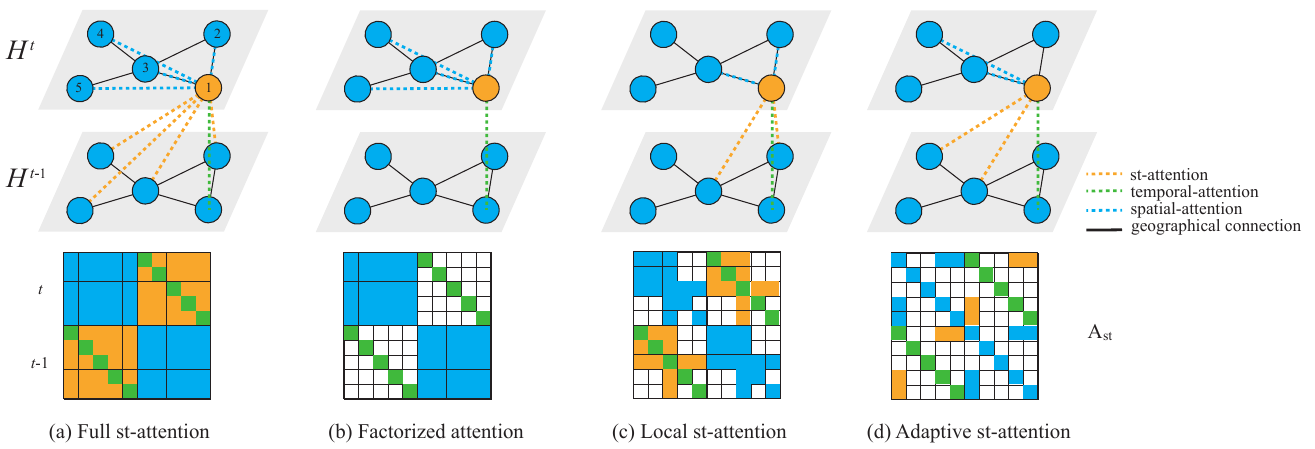}
  \caption{Different kinds of attentions can be considered to build spatial-temporal graph transformer. The upper panel denotes the same graph (with self-connection) at two time steps while node $1$ at time $t$ wants to update its own represention based on other possibly related nodes using attention mechanism. The lower panel denotes the corresponding spatial-temporal attention matrix. Colored entry means the corresponding two nodes (determined by row and coloumn indices) will attend to each other. Different colors correspond to different types of attentions defined in the caption. }
  \label{fig:attention}
\end{figure*}

\subsection{Local Spatial-Temporal Attention}
\label{sec:lsta}
Dealing with spatial-temporal data requires modeling the correlations in both spatial and temporal domains. We refer to attentions between nodes with the same time step (location) as spatial-(temporal-)attention, and use st-attention to denote attentions for node pairs with both different time steps and locations.
As shown in Figure \ref{fig:attention}(a), the most straightforward way to fuse the spatial and temporal attention is to perform attention on every pair of nodes in the spatial-temporal graph (full st-attention) and treat each individual node representation $h_{t,i}\in\mathbb{R}^{D}$ as a token (with the input node feature matrix $H=[h_{t,i}]_{ti}\in\mathbb{R}^{T\times N\times D}$), and $h_{t,i}$ can attend over all tokens from the input. Although this approach can capture attention of two nodes which may be far away from each other both in spatial and temporal domains, it induces a complexity of $\mathcal{O}(T^2N^2)$ which cannot scales to large graph as $N$ goes larger.

An alternative approach to reduce the time complexity is to factorize st-attention to spatial and temporal dimension separately, then calculate the attention one by one \cite{zheng2020gman}. Concretely, as shown in Figure \ref{fig:attention}(b), spatially, each node can only horizontally attend over nodes with the same time step, and temporally, each node can only vertically attend over nodes at the same location. Different heads are used to capture spatial and temporal correlations. This approach has a reduced time complexity of $\mathcal{O}(T^2+N^2)$, but it does not directly consider the dynamic st-attention and obtains st-attention by combining the spatial-attention and temporal-attention.

To efficiently capture the node correlations in both spatial and temporal dimensions, we apply local spatial-temporal multi-head attention for node update. As shown in Figure \ref{fig:attention}(c), we use the spatial adjacency to reduce the complexity of attentions by keeping the st-attention within the spatial 1-hop neighborhood. Concretely, we limit the scope of each node to its geographical neighbors and calculate attentions over all nodes with these geographical positions. Nodewisely, it can be written as
\begin{equation}
\label{eq:local_att}
\begin{split}
   h_{t,i} = \sum_{j\in \mathcal{N}(i)\cup \{ i\}} \sum_{t'}\  \alpha_{t't,ji} v_{t',j},
\end{split}
\end{equation}
where $\mathcal{N}(i)$ means the neighborhood multiset of node $i$, $\alpha_{t't,ji}$ means attention between node $(t,i)$ and $(t',j)$, and $v_{t',j}$ means the projected value based on $h_{t',j}$. In the matrix form (like in \ref{sec:msa}), this is equivalent to applying appropriate st-attention mask to the full st-attention matrix (shown in Figure \ref{fig:attention}(b) lower panel). Specifically, we can flatten the input feature matrix to $H\in\mathbb{R}^{TN\times D}$, and the st-attention mask is denoted by $A_{st}\in\mathbb{R}^{TN\times TN}$. Then the local multi-head self-attention (L-MSA) can be written as
\begin{equation}
\label{eq:lsa}
\begin{split}
   L{\text -}SA(X,A_{st}) = \left[softmax\left(\frac{QK^T}{\sqrt{D_{QK}}}\right)\cdot A_{st}\right]V,
\end{split}
\end{equation}
\begin{equation}
\label{eq:lmsa}
\begin{split}
    L{\text -}MSA(X,A_{st}) = concat(L{\text -}SA(X,A_{st})_1,...,L{\text -}SA(X,A_{st})_h)W_O.
\end{split}
\end{equation}
Restricting attentions to spatial neighborhood is reasonable because the long-range spatial correlations can be captured as the number of layers increases, and the long-range st-attentions are usually weak. We keep the full attention along time dimension because $T$ is usually small compared to the number of different locations $N$. The corresponding time complexity is $\mathcal{O}(ET^2)$, which is scalable especially for sparse graph. $E$ here represents the total number of edges in the spatial graph. 

Furthermore, motivated by the fact that the geographical adjacency may not reflect the genuine dependency relationships among nodes \cite{wu2019graph}, we further improve the local spatial-temporal attention by introducing the adaptive learnable adjacency matrix $A_{apt}$, which does not require any prior knowledge and can be learned end-to-end. We use two learnable node embeddings $U_1, U_2 \in \mathbb{R}^{N\times c}$ with random initialization to calculate $A_{apt}$ as 
\begin{equation}
\label{eq:apt}
\begin{split}
   A_{apt} = softmax(U_1U_2^T).
\end{split}
\end{equation}
However, the calculated adjacency matrix will induce a complete graph (where every pair of nodes is connected), and applying local attentions on it will be equivalent to the full st-attention approach. Therefore, we calculate the binary mask $b$ using Gumbel-sigmoid trick \cite{jang2016categorical} based on every entry of $A_{apt}$ and set the maximum in-degree (maximum number of non-zero values of each row), then apply it to $A_{apt}$ with elementwise multiplication
\begin{equation}
\label{eq:apt}
\begin{split}
   A_{apt} = b\cdot A_{apt}.
\end{split}
\end{equation}

\subsection{Framework of ASTTN}
\begin{figure}[t]
  \centering \includegraphics[width=0.98\linewidth]{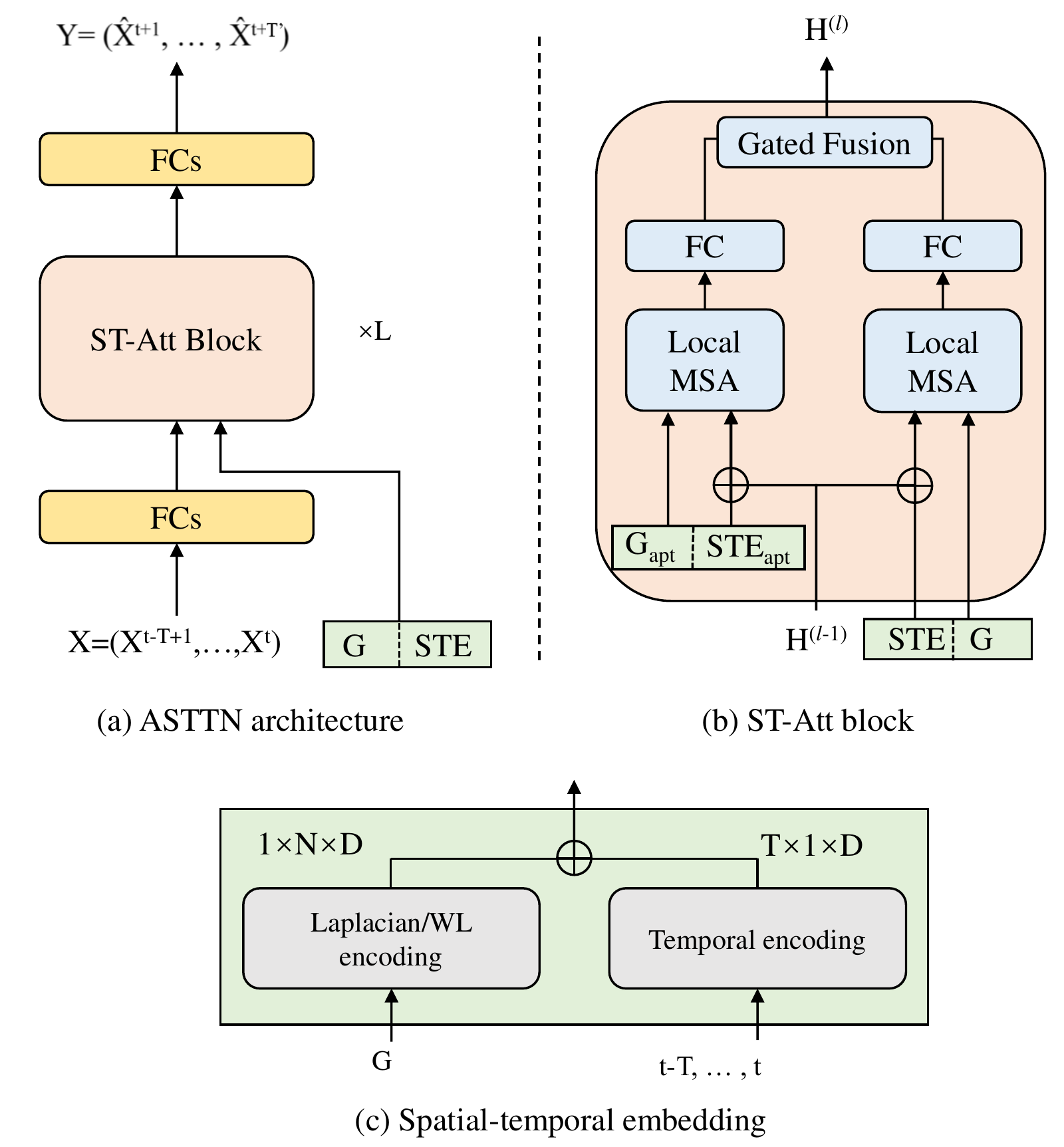}
  \caption{The framework of Adaptive Spatial-Temporal Transformer Networks (ASTTN). (a) ASTTN takes node feature matrix together with the graph structure and spatial-temporal embeddings as input. It consists of fully connected input and output layers, stacked ST-Att Blocks. (b) Each ST-Att Block performs two local multi-head self-attention (local MSA) on both input graph and trainable adaptive graphs, and fuse the results to obtain outputs. (c) Spatial embedding and temporal embedding are calculated separately, and then projected to the same dimension with linear layers. The spatial-temporal embedding is obtained by summing up spatial and temporal embedding through tensor broadcasting.}
  \label{fig:archi}
\end{figure}

We present the framework of ASTTN in Figure \ref{fig:archi}. It consists of input layer, spatial-temporal embedding layer, stacked st-attention blocks with residual connections \cite{he2016deep} and output layer. The input to the model includes a feature matrix $X\in\mathbb{R}^{T\times N\times D_{in}}$ and an underlying graph structure $G$. The input and output of each st-attention block are denoted as $H^{(l-1)}$ and $H^{(l)}\in\mathbb{R}^{T\times N\times D}$, which have the same dimension $D$ to facilitate residual connections. The graph structures are utilized in embedding layer to encode the structural information and st-attention block to calculate local MSA. Detailed module illustrations are given below.

\subsubsection{Spatial-Temporal Embedding}
\label{sec:ste}
Since the transformer-based model contains no recurrence and no convolution, it is important to add positional encoding to each input token to ensure unique representation and preserve the distance information. For graph structure, the design of such positional encoding remains open questions \cite{dwivedi2021graph,dwivedi2020benchmarking} because of permutation-invariant nature of GNNs. We consider the Laplacian positional encoding for graph structure as used in \cite{dwivedi2020generalization}. Specifically, we compute the laplacian eigenvectors of the input graph as
\begin{equation}
\label{eq:L}
\begin{split}
   L= T-D^{-1/2}AD^{-1/2}=U^T\Lambda U,
\end{split}
\end{equation}
where $A$ is the adjacency matrix, $\Lambda$ is the eigenvalue matrix and $U$ is the eigenvectors matrix. Then we use the $k$ smallest non-trivial eigenvectors of a node as its positional embedding. The positional embeddings are then fed into FC layer to keep the same dimension as the input $X$.
For the sequential temporal connections, we simply utilize the time step to generate temporal embeddings, considering the periodicity of the traffic, we follow \cite{zheng2020gman} and use the day-of-week and time-of-day of each time step to compose the two-dimensional temporal encoding, followed by FC layer. Then we sum up the positional embedding and and temporal embedding to obtain the st-embedding (STE) which describes the unique position of one node in the graph structure across different time steps.

\subsubsection{ST-Attention Block}
ST-attention block performs local spatial-temporal attention (discussed in \ref{sec:lsta}) on the input traffic matrix $H^{(l-1)}\in\mathbb{R}^{N\times T\times D}$. As shown in Figure \ref{fig:archi}(b), it consists of two parallel local multi-head attention modules based on two different sets of graphs and st-embeddings. Two types of graph structures are considered in this block. The first is the original road map graph $G$ whose adjacency matrix $A\in\mathbb{R}^{N\times N}$ is determined by the original geographical connections. The second is the adaptive graph $G_{apt}$ whose adjacency matrix $A_{apt}$ is totally parameterized as in Eq. \ref{eq:apt}. Two types of spatial-temporal embeddings STE and ${\text{STE}}_{\text{apt}}$ are calculated based on $G$ and $G_{apt}$, respectively, as described in \ref{sec:ste}, and then added up to the input $H^{(l-1)}$. Next, local MSA performs local spatial-temporal attention following Eq. \ref{eq:lsa} and \ref{eq:lmsa}, where the input st-feature matrix is the flattened input $H^{(l-1)}$ and the input st-attention mask is constructed from $A$ or $A_{apt}$ as shown in Figure \ref{fig:attention} lower panel. It should be noted that in the implementation, we do not explicitly construct such st-feature matrix and st-attention mask which scale as $T^2N^2$, but instead use more memory-efficient and equivalent calculations discussed in the next section. The outputs from the parallel local MSA modules will then be fused together using gated fusion mechanism as used in \cite{wu2019graph}. Compared to previous works which use separated modules for spatial and temporal domain modeling, our st-attention module can update node embedding in both spatial and temporal domains simultaneously.

\subsubsection{Input and Output Layers}
The input and output layers are fully-connected networks (FC) with ReLU activation. The input layer is used to map the input node feature to a higher dimension $D$. The output layer is used to map the temporal dimension from history time steps $T$ to the future temporal dimension $T'$. The prediction $\hat{Y}=[\hat{X}_{t+1},...,\hat{X}_{t+T'}]$ is then used to calculate the mean absolute error (MAE) with respect to the ground truth $Y=[X_{t+1},...,X_{t+T'}]$:
\begin{equation}
\label{eq:loss}
\begin{split}
   L = \frac{1}{T'}\sum_{t'=t+1}^{t'=t+T'}|\hat{X}_{t'}-X_{t'}|,
\end{split}
\end{equation}
which is used to train ASTTN end-to-end through back-propagation.

\section{Experiments}

\begin{table*}[h!]
	\centering
	\begin{tabular}{l l  r r r | r r r |r r r | r r r}
		\toprule
		\multirow{2}{*}{Data} & \multirow{2}{*}{Models}   & \multicolumn{3}{c}{15 min} & \multicolumn{3}{c}{30 min}  & \multicolumn{3}{c}{60 min}  & \multicolumn{3}{c}{120 min}\\ 
		\cline{3-5}  \cline{6-8} \cline{9-11}  \cline{12-14}  
		&& {\small MAE} & {\small RMSE} & {\small MAPE} & {\small MAE} & {\small RMSE} & {\small MAPE} & {\small MAE} & {\small RMSE} & {\small MAPE} & {\small MAE} & {\small RMSE} & {\small MAPE}\\
		\midrule
		\multirow{9}{*}{\rotatebox[origin=c]{90}{METR-LA}}
		&ARIMA         & 3.99 & 8.21 & 9.60\% & 5.15 & 10.45 & 12.70\%& 6.90 & 13.23 & 17.40\% & 9.32 & 15.67 & 20.64\% \\
		&LSVR          & 3.99 & 8.45 & 9.30\% & 5.05 & 10.87 & 12.10\%& 6.72 & 13.76 & 16.70\% & 8.94 & 13.85 & 18.15\% \\
		&FC-LSTM       & 3.44 & 6.30 & 8.60\% & 3.77 & 7.23 & 10.90\% & 4.37 & 8.69 & 13.20\%  & 6.53 & 10.61 & 14.45\% \\
		&DCRNN         & 2.77 & 5.38 & 7.30\% & 3.15 & 6.45 & 8.80\%  & 3.60 & 7.60 & 10.50\%  & 6.34 & 10.28 & 14.34\% \\ 
		&STGCN         & 2.88 & 5.74 & 7.62\% & 3.47 & 7.24 & 9.57\%  & 4.59 & 9.40 & 12.70\%  & 6.82 & 11.21 & 15.26\% \\
	    &ASTGCN        & 2.75 & 5.62 & 7.51\% & 3.31 & 6.98 & 9.32\%  & 4.52 & 9.24 & 12.62\%  & 6.12 & 9.76 & 13.65\% \\
	    &Graph WaveNet & \textbf{2.69} & \textbf{5.15} & \textbf{6.90\%} & \textbf{3.07} & 6.22 & \textbf{8.37\%}  & 3.53 & 7.37 & 10.01\% & 5.82 & 8.53 & 13.07\% \\
	    &GMAN          & \textbf{2.69} & 5.55 & 7.42\% & 3.15 & 6.78 & 9.02\%  & 4.03 & 8.11 & 11.72\%  & 5.42 & 8.64 & 12.92\% \\
		&ASTTN (Ours)  & 2.74 & 5.45 & 7.48\% & 3.10 & \textbf{6.15} & 8.53\% &  \textbf{3.13} & \textbf{7.10} & \textbf{9.84}\%   & \textbf{5.20} & \textbf{8.28} & \textbf{12.75\%} \\
		\midrule
		\multirow{9}{*}{\rotatebox[origin=c]{90}{PEMS-BAY}}
		&ARIMA         & 1.62 & 3.30 & 3.50\% & 2.33 & 4.76 & 5.40\%  & 3.38 & 6.50 & 8.30\%  & 5.13 & 8.04 & 11.87\% \\ 
		&LSVR          & 1.85 & 3.59 & 3.80\% & 2.48 & 5.18 & 5.50\%  & 3.28 & 7.08 & 8.12\%  & 4.97 & 7.76 & 11.45\% \\ 
		&FC-LSTM       & 2.05 & 4.19 & 4.80\% & 2.20 & 4.55 & 5.20\%  & 2.37 & 4.96 & 5.70\%  & 4.63 & 7.12 & 10.38\% \\ 
		&DCRNN         & 1.38 & 2.95 & 2.90\% & 1.74 & 3.97 & 3.90\%  & 2.07 & 4.74 & 4.90\%  & 4.52 & 7.15 & 10.08\% \\ 
		&STGCN         & 1.36 & 2.96 & 2.90\% & 1.81 & 4.27 & 4.17\%  & 2.49 & 5.69 & 5.79\%  & 4.72 & 7.31 & 10.62\% \\ 
	    &ASTGCN        & 1.32 & 2.78 & 2.75\% & 1.75 & 3.98 & 3.95\%  & 2.32 & 5.41 & 5.51\%  & 4.42 & 7.02 & 9.76\%  \\ 
	    &Graph WaveNet &\textbf{1.30}&2.74&\textbf{2.73\%}  & 1.63&\textbf{3.70}&3.67\% & 1.95 & 4.52 & 4.63\%       & 4.27 & 6.95 & 9.05\% \\ 
		&GMAN          & 1.34 & 2.82 & 2.81\% & 1.62 & 3.72 & \textbf{3.63\%}  & 1.86 & 4.32 & 4.31\%  & 3.95 & 6.61 & 8.89\% \\ 
		&ASTTN (Ours)  & 1.32 & \textbf{2.70} & 2.78\% & \textbf{1.58} & 3.72 & 3.64\%  & \textbf{1.72} & \textbf{4.02} & \textbf{3.98\%}  & \textbf{3.82} & \textbf{6.33} & \textbf{8.68\%} \\
		\bottomrule
	\end{tabular}
	\caption{Performance comparison of ASTTN and other baseline models. ASTTN always achieves the best or second best results on both datasets from short-term to long-term prediction.} 
	\label{table1:metr}
\end{table*}
In this section, we present the experimental results of ASTTN and competing baselines over four spatial-temporal traffic datasets, i.e., METRA-LA, PEMS-BAY released by Li et al. \cite{li2017diffusion}, and PeMSD4, PeMSD7 processed by Guo et al. \cite{guo2019attention}.
We also analyze the model performance with different types of attention and model configurations for ablation study. 
\subsection{Datasets}
The dataset descriptions are as follows:
\begin{itemize}
  \item \textbf{METR-LA} records 4 months of traffic speed data on 207 sensors ranging from Mar 1st 2012 to Jun 30th 2012, collected from loop detectors in the highway of Los Angeles County.
  
  \item \textbf{PEMS-BAY} contains 6-month traffic information on 325 sensors ranging from Jan 1st 2017 to May 31th 2017, collected by California Transportation Agencies (CalTrans) Performance Measurement System (PeMS).
  
 \item \textbf{PeMSD4} is collected in San Francisco Bay Area with 29 roads from January to February in 2018, containing 307 detectors.
 
  \item \textbf{PeMSD7} is collected in San Francisco Bay Area from May through June in 2012, containing 228 detectors.

\end{itemize}

\subsection{Experimental Setting}
Our experiments are conducted on a computer with Intel(R) Core(TM) i9-10920X CPU @ 3.50GHz CPU with NVIDIA GPU (GeForce RTX 3090). Following previous works, we use $T=12$ (1 hour) as the historical time steps to predict the future traffic condition of the next $T'=3, 6, 12, 24$ steps.
We train our model using Adam optimizer with the initial learning rate 0.001.
To construct the adaptive graph, we create random-initialized node embeddings $U_1, U_2$ following uniform distribution with dimension size 10.
The hyper-parameters to be tuned in the model include the number of st-attention block $L$, the number of attention heads $K$, the dimension of each head $d$ (intermediate node dimension $D=d\times K$), the in-degree threshold to mask the adaptive graph.
We adopt three widely used metrics to evaluate the traffic prediction performance, i.e., Mean Absolute Error (MAE), Root Mean Squared Error (RMSE), and Mean Absolute Percentage Error (MAPE). The model is trained on training dataset, the model selection is performed on validation dataset, and the performance is reported on test dataset.

\subsection{Implementation Details}
We implement the model with PyTorch \cite{NEURIPS2019_9015}, and to leverage the sparsity of input graph, we utilize DGL \cite{wang2019deep} package which can perform fast and memory-efficient message passing primitives for training graph neural networks. To efficiently calculate the st-attention, we only build the spatial graph using DGL to avoid large memory usage (compared to building the whole spatial-temporal graph) and calculate $Q, K, V\in\mathbb{R}^{T\times N\times D}$ with Eq. \ref{eq:qkv}. Next, we fix $Q$ matrix, while rolling $K, V$ matrix by 1 along the first temporal dimension, and assign this $Q, K, V$ pair to each node to calculate attention with Eq. \ref{eq:sa}, using messaging passing enabled by DGL. In this way, we are actually calculating the st-attention between two adjacent time steps (orange-dashed line in Figure \ref{fig:attention}). We repeat this rolling process for $T$ times, and summing up these results is equivalent to calculating Eq. \ref{eq:lsa}.

\begin{table}[bp!]
\centering
\begin{tabular}{lllll}
\hline 
Data& Models         & MAE & RMSE & MAPE \\ \hline
\multirow{6}{*}{\rotatebox[origin=c]{90}{PeMSD4}}
&DCRNN          & 4.15 & 8.20 & 10.82\% \\ 
&STGCN          & 4.08 & 7.69 & 10.23\% \\
&ASTGCN         & 3.96 & 7.20 & 10.53\% \\
&Graph WaveNet  & 3.75 & 7.02 & 9.58\% \\
&GMAN           & 3.78 & 7.10 & 9.72\% \\
&ASTTN (Ours)   & \textbf{3.52} & \textbf{6.88} & \textbf{9.54\%} \\
\midrule
\multirow{6}{*}{\rotatebox[origin=c]{90}{PeMSD7}}
&DCRNN          & 2.26 & 5.28 & 5.10\% \\ 
&STGCN          & 2.55 & 5.65 & 5.39\% \\
&ASTGCN         & 2.73 & 5.21 & 5.46\% \\
&Graph WaveNet  & 2.03 & 4.65 & 4.60\% \\
&GMAN           & 2.05 & 4.60 & 4.52\% \\
&ASTTN (Ours)   & \textbf{1.92} & \textbf{4.43} & \textbf{4.36}\% \\
\hline
\end{tabular}
\caption{Performance comparison of ASTTN and other baseline models on PeMSD4 and PeMSD7 dataset for 60 minutes ahead prediction.}
\label{table2:pemsd}
\end{table}

\begin{table*}[ht!]
\centering
\begin{tabular}{llllll}
\hline 
Type    & MAE & RMSE & MAPE & Time (ms) & Complexity\\ \hline
Full st-attention      & 1.93 & 4.53 & 5.05 \%  & 936.4 & $\mathcal{O}(T^2N^2)$\\
Factorized attention   & 1.85 & 4.34 & 4.63 \%  & 112.3 & $\mathcal{O}(T^2+N^2)$ \\
Local st-attention     & 1.83 & 4.26 & 4.24 \%  & 120.6 &$\mathcal{O}(ET^2)$\\
Adaptive st-attention  & 1.74 & 4.16 & 4.45\%   & 139.5 &$\mathcal{O}(ET^2)$\\ 
Local+Adaptive (ours)  & 1.72 & 4.02 & 3.98\%   & 206.1 & $\mathcal{O}(ET^2)$ \\ 
\hline
\end{tabular}
\caption{Comparison of performance on PEMS-BAY (60 min) dataset for different types of attentions as shown in Figure \ref{fig:attention}.}
\label{table3:attention}
\end{table*}
\subsection{Baselines}
We compare the performance of ASTTN with the following spatial-temporal modeling benchmarks:
\begin{itemize}
    \item ARIMA \cite{williams2003modeling} is the classic time series forecasting method using the correlation between time series data to predict traffic.
    
    \item LSVR \cite{wu2004travel} refers to using support vector machine for future traffic prediciton.
    
    \item FC-LSTM \cite{sutskever2014sequence} is a sequence-to-sequence model with fully-connected LSTM layers in both encoders and decoders.
    
    \item DCRNN \cite{li2017diffusion} refers to Diffusion Convolutional Recurrent Neural Network which incorporates both spatial and temporal dependencies into a sequence-to-sequence framework for traffic flow prediction.
    
    \item STGCN \cite{yu2017spatio} builds the structure with complete convolutional modules using graph convolution for spatial domain and 1-D convolution for temporal domain.
   
    \item ASTGCN \cite{guo2019attention} combines the spatial-temporal attention mechanism and simultaneously captures the dynamic spatial-temporal characteristics of traffic data using convolutions.
    
    \item Graph Wavenet \cite{wu2019graph} introduces a novel adaptive dependency matrix which can be learned through node embeddings.

    \item GMAN \cite{zheng2020gman} adopts an encoder-decoder architecture. Both the encoder and the decoder are composed of multiple spatial-temporal attention modules, and the gate fusion mechanism is used to merge the influence of spatial-temporal factors on traffic information.
\end{itemize}

\subsection{Experimental Results}
\begin{figure}[hb]
  \centering \includegraphics[width=0.98\linewidth]{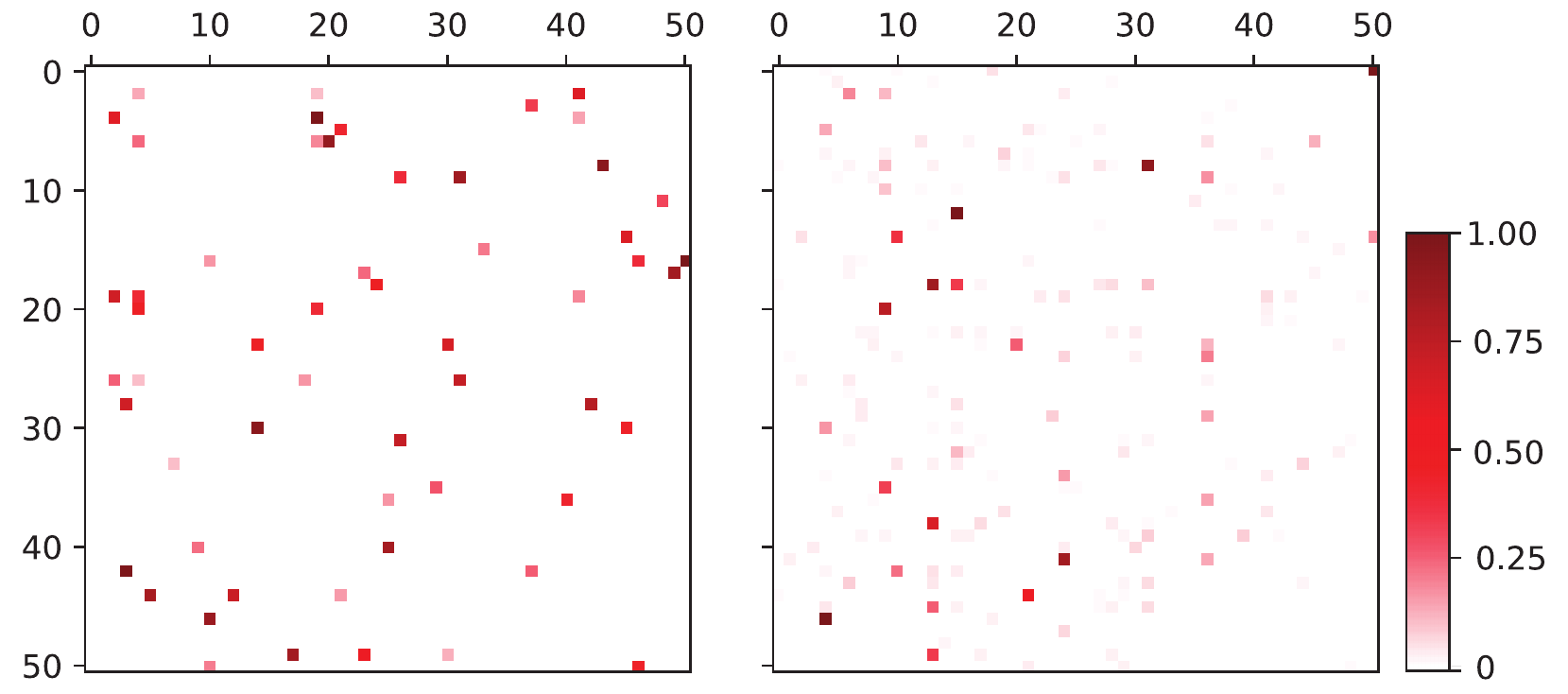}
  \caption{Comparison of the input geographical adjacency matrix (left) and the learned adaptive attention matrix (right) for the first 50 nodes in PeMS-BAY dataset.}
  \label{fig:apt}
\end{figure}

We compare the performance of ASTTN with benchmark models in Table \ref{table1:metr} for 15 minutes (3 steps), 30 minutes (6 steps), 60 minutes (12 steps) and 120 minutes (24 steps) on
METR-LA and PEMS-BAY datasets. We find that (1) our model along with other deep-learning-based models which consider graph structures are able to outperform conventional methods ARIMA, LSVR, and FC-LSTM, which shows the power of deep learning model and the importance of incorporating the graph structure into temporal prediction. (2) Graph WaveNt and our model outperform graph models DCRNN and STGCN for graph structure modeling, indicating that the genuine spatial dependencies need to be explored to improve the model performance. (3) GMAN and our model also outperform conventional graph deep learning models, which shows the importance of capturing the dynamic spatial-temporal correlations. (4) Our model achieves state-of-the-art prediction performance compared to benchmarks, and the advantages are more evident in the long-term horizon prediction. The relative low performance in short-term prediction can be possibly explained by the fact that the st-attention effect is relatively weak while spatial-attention can play a more important role because the changes in the temporal dimension are relatively small. We confirm the superior performance of our model on another two datasets PeMSD4 and PeMSD7 regarding long-term prediction, and as shown in Table \ref{table2:pemsd}, our model outperforms the benchmark models on both datasets. 

\begin{table}[ht!]
\centering
\begin{tabular}{llll}
\hline 
model         & 15min & 30min & 60min \\   \hline
ASTTN-NE      & 1.43 & 1.66 & 1.84 \\
ASTTN-NF      & 1.38 & 1.60 & 1.78  \\
ASTTN-NA      &1.45 & 1.69 & 1.82  \\
ASTTN         & 1.32 & 1.58 & 1.72 \\ 
\hline
\end{tabular}
\caption{Comparisons of model variants' MAE on PEMS-BAY.}
\label{table4:variant}
\end{table}

To investigate the effect of each component in our model, we then evaluate the model variants' performance by removing the spatial-temporal node embedding (ASTTN-NE), gated fusion (ASTTN-NF), and adaptive local-MSA module (ASTTN-NA). As shown in Table \ref{table4:variant}, ASTTN consistently outperforms its variants, which indicates the importance of the spatial-temporal embedding, gated fusion, and adaptive local-MSA in capturing the complex spatial-temporal dependencies. 

\subsection{Effect of spatial-temporal attention}

We show the effect of different types of spatial-temporal attention as discussed in Section \ref{sec:lsta}. To make fair comparisons, we fix the hyper-parameters and only change the attention mechanisms (Figure \ref{fig:attention}) of the local MSA modules. We show the comparisons in Table \ref{table3:attention}. The running time refers to the averaged forward pass time with the same input batch size. We can see that full st-attention is hard to train and performs the worst potentially because of being over-fitted to the dataset. Factorized attention performs better than full st-attention and run fastest. Local and adaptive st-attention have comparable running time compared with factorized attention, while adaptive st-attention performs better than the first three methods, because it can explore the genuine node interactions. Finally, the combined local and adaptive attention used in our paper achieves the best prediction performance with also acceptable running time.

We further investigate the learned adaptive adjacency matrix with PEMS-BAY datasets. As shown in Figure \ref{fig:apt}, the learned adaptive adjacency shows more sparse correlations between nodes compared with the original geographical adjacency matrix because we apply the mask based on the correlation values. Besides, the adaptive adjacency matrix reveals many correlations that are not shown in the input adjacency graph, which means the geographical correlations may not be able to describe the real node dependencies.

\section{Conclusion}
In this paper, we propose a novel model termed ASTNN for spatial-temporal modeling with the graph structure for traffic prediction. ASTNN is built from stacked st-attention blocks which are used to simultaneously model the spatial and temporal correlations. We use local multi-head self-attention to efficiently calculate attentions on the spatial-temporal graph. Besides, to explore the genuine spatial correlations and improve the local spatial-temporal attention's performance, we introduce the learnable adaptive graph which can help the target node to select related nodes to attend over. We compare the effect of different types of spatial-temporal attention and show the effectiveness of local spatial-temporal attention. Comprehensive empirical studies on four traffic datasets show the superior performance of ASTNN compared with state-of-the-art benchmarks. Ablation studies and visualization of adaptive adjacency matrix show the influence of each component of our model. In the future work, We plan to investigate the influence of different types of attention on more complex spatial-temporal graph, e.g., with dynamically changing topology.
\bibliographystyle{ACM-Reference-Format}
\bibliography{acmart}


\end{document}